\documentclass[conference]{IEEEtran}
\IEEEoverridecommandlockouts
\usepackage{cite}
\usepackage{amsmath,amssymb,amsfonts}
\usepackage{algorithmic}
\usepackage{graphicx}
\usepackage{textcomp}
\usepackage{xcolor}
\usepackage{booktabs}
\usepackage{tabularx}
\usepackage{hyperref}

\hypersetup{
    colorlinks=true,
    linkcolor=black,
    citecolor=blue,
    urlcolor=blue,
}

\def\BibTeX{{\rm B\kern-.05em{\sc i\kern-.025em b}\kern-.08em
    T\kern-.1667em\lower.7ex\hbox{E}\kern-.125emX}}

\begin{document}

\title{Real-Time Automatic License Plate Recognition Using YOLOv8, SORT Tracking, and Temporal Data Interpolation\\
}

\author{\IEEEauthorblockN{Mirza Muhammad Mobeen}
\IEEEauthorblockA{{Software Engineer}, Sanwa Comtec K.K. Japan \\
{Researcher}, National University of Technology (NUTECH), Pakistan}
}

\maketitle

\begin{abstract}
The real-time hardships of video processing seriously limit the usage of Automatic License Plate Recognition (ALPR) with application in dynamic traffic monitoring settings. High-fidelity recognition of unconstrained variables, e.g. drastic variations in illumination, acute camera scans, high vehicle speeds, and harsh physical concealment, is a problem that often leads to disjointed tracking paths and poor Optical Character Recognition (OCR) rates. In order to mitigate these weaknesses, the study proposes a 5 stage, end-to-end algorithmic pipeline, encompassing a smooth transition between deep learning based object detection, multi-object tracking which is kinematic in nature, and geometry temporal data interpolation. The suggested architecture takes advantage of a very powerful YOLOv8 nano model to localize the vehicle at the first stage and then Simple Online and Realtime Tracking (SORT) algorithm is used to build spatial-temporal links between frames. Another, more specific typology of YOLOv8 object detectors the license plate area, channeling the sliced array to an EasyOCR chain under the limitations of positional syntax verification. More importantly, an offline interpolation mechanism of temporal bounding box is initiated to recast fragmented paths. Operational efficacy of the system has been proven as shown in empirical assessment performed on 3,599 continuous frames of traffic surveillance footage. The pipeline was able to track 45 different vehicles yielding 2,247 raw baseline detection. Algorithms with the linear temporal interpolation imputed a total of 2,289 missing spatial entries and thus produced the ultimate enriched version of a dataset consisting of 4,536 coordinates, an absolute increment of 101.9\%. Although its top-score was reached at 0.982 on the OCR confidence scale, the average OCR confidence was 0.414, continuing to demonstrate the unstable image-quality performance of working with frames. The main contributions within this piece of work are the new combination of YOLOv8, SORT, and EasyOCR based on the original work, coupled with the detailed addition of an efficient temporal interpolation layer that effectively eliminates the negative effect of occlusion on the data loss in end-to-end deployments of ALPR.
\end{abstract}


\section{Introduction}
The blistering development of urban infrastructure and the corresponding increase in the number of vehicles around the world has dictated the introduction of Intelligent Transportation Systems (ITS) on a large scale \cite{jain2019review}. At the core of the operational performance of contemporary ITS is Automatic License Plate Recognition (ALPR) which is a computer vision technology that effectively makes the electronic toll collection, automated parking control, law enforcement, and wide-area surveillance security operational. Whereas ALPR systems have excellent performance in highly regulated, stationary situations, like toll readers mounted on gantries with special infrared lighting, identifying alpha numeric characters in unstructured, moving video unaided is an incredibly difficult technological endeavor.

Surveillance cameras installed in the real world have to consume video streams that are exposed to severe environmental stochastics \cite{du2013automatic}. Lighting conditions that vary, such as direct sun glare, structural shadows, poor-quality light sensors in low light, etc., will continuously modify the character shape, which will need reference to a different contrast profile to ensure proper character segmentation. Moreover, cars are moving faster across the field of view of the camera and over a wide range of patterns that cause motion lack, acute angular distortions, which destroy pixel-level faithfulness. Due to heavy urban traffic, the line of sight is often truncated by the following cars, pedestrians, or stationary infrastructure resulting in disastrous tracking losses. The generic recognition algorithms are further complicated by the inherent variety of the license plates themselves; that is, the difference in aspect ratios, regional syntax systems, hand centered fonts, and physical corruption.

In order to avoid the very nature of the limitations posed by frame-by-frame analysis, this study proposes an effective, tracking-by-detection architecture which is aimed at imposing temporal consistency within the sequence of the video. The transient OCR failures and spatial dropouts can be reduced by binding together individual, frame-level targets with long term object identities using an algorithmic process.

These are the contributions of the particular research:
\begin{itemize}
    \item The architecture, training, and deployment of a very efficient, end-to-end ALPR framework combining the anchor-free YOLOv8 structure to detect, the SORT algorithm to track several objects, and the CRNN based EasyOCR system to extract text.
    \item Thiscan be achieved through an interpolation algorithm that uses a temporal bounding box algorithm that is based on the use of linear geometric imputation to seamlessly repurpose dropped tracks to actually increase continuous tracking of positional information by 101.9\%.
    \item The creation of a strong, syntax-conscious OCR follow-on-processing unit on the UK license plate format, with a positional character-correction mapping table to actively remove misclassifications of common alpha-numeric neural networks.
    \item Full empirical analysis of the combined pipeline on a continuous 3599-frame traffic monitoring sequence, which gives a quantitative analysis of the intricate association between confirmation of the detection, geometry of the bounding box, splitting of the tracking, and stability of the OCR grammar.
\end{itemize}

Further format of this paper is organized as follows: Section II reviews related work. Section III details the mathematical and architectural framework of the proposed methodology. Section IV outlines the experimental setup. Section V presents quantitative results. Section VI provides a critical discussion of the empirical findings and system limitations. Finally, Section VII highlights data availability, and Section VIII concludes the research.

\section{Related Work}
The continuous evolution of computer vision has fundamentally transformed the architectural design of ALPR systems, shifting the paradigm from deterministic image processing to highly parameterized deep learning networks.

\subsection{Object Detection Evolution}
A lifelong game has been played on the path of object detection, where there is a relentless attempt to compromise between spatial accuracy (localization) and temporal latency (computation time). Earlier methods made use of R-CNN and Faster R-CNN \cite{ren2017faster}, which were highly accurate with the use of Region Proposal Networks however had the high cost of computation. This was resolved by single-stage detectors such as SSD has shown to detect objects at once: the SSD detector has fixed some flaws by extending the frame for one unified objective, another one is the YOLO family series, which includes the You Only Look Once network: a series of network types that redefine object detection as an uninterrupted regression problem. Another state of the art was proposed by YOLOv8 \cite{jocher2023ultralytics} that offered an anchor-free head, therefore directly predicting the center of objects. Integration of C2f module augmented gradient flow, and loose-coupled-head divided objectness, class and regress activity set a brand new standard, real-time inference of edges, a state of the art.

\subsection{License Plate Detection and Recognition}
The classical approach would have massively depended on heuristics such as Canny edge detector and morphological dilation to perform this task touch-on-feelingly \cite{du2013automatic}. Although these were computationally lightweight, they had deplorable brittleness. These were substituted by powerful CNNs with a tendency to generalize under diverse conditions with the evolution of deep learning studies \cite{silva2018license, li2019toward, xie2018new}.

\subsection{Multi-Object Tracking (MOT)}
It is not enough to identify an object and a single frame, but the system should ensure its temporal continuity is established \cite{luo2021multiple}. A Kalman filter (used for linear state prediction)\cite{kalman1960new} with Hungarian algorithm \cite{kuhn1955hungarian}(used for data association) based on Inclusion of Bounding Box IoU is used in the Simple Online and Realtime Tracking (SORT) algorithm\cite{bewley2016simple}. Although extremely vulnerable to identity (ID) switches even when AC are long, SORT is still used in resource constrained ITS deployments since it is faster. These algorithms were followed by later counterparts such as DeepSORT\cite{wojke2017simple} , best in tracking sorts usable for violently head-on collisions - with deep appearance embeddings, or legality tracking with report-and-follow deep organ views, like bytetrack\cite{zhang2022bytetrack} .

\subsection{Optical Character Recognition (OCR)}
With old-fashioned engines (as Tesseract did)\cite{smith2007overview}  like this, it works well on scanned paperwork but fails in-the-field. More contemporary methods like EasyOCR\cite{jaided2020easyocr}, which uses Convolutional Recurrent Neural Networks (CRNNs) based on ResNet backbones, Bidirectional LSTM, and Connectionist Temporal Classification (CTC), can use alignments without the explicit character-scale segmentation required in more traditional methods  \cite{shi2017end}.

\subsection{Gap Analysis}
The dropouts of frame-level detection is an area where there is an acute gap in terms of the systemic treatment. Statistical approaches to temporal data completion have tended to use post-processing to complete the data such as linear interpolation \cite{virtanen2020scipy}. Nevertheless, the currently employed methodologies often do not have an inbuilt time interpolation component, which acts as an end-to-end feedback mechanism on end-to-end ALPR pipelines.This research directly addresses this gap.

\section{Proposed Methodology}
The proposed architecture is structured as a sequential, five-stage pipeline designed to ingest raw video frames, detect and track vehicles, isolate license plate geometries, extract alphanumeric text, and algorithmically impute missing spatial data.

\subsection{Stage 1: Vehicle Detection (YOLOv8n)}
The primary detection module employs the nano variant of the Ultralytics YOLOv8 architecture (YOLOv8n). Selected explicitly for its extreme parameter efficiency (3.2 million parameters, 6.5 MB physical size), the model is highly optimized for real-time inference on edge hardware. 

Operating as an anchor-free detector, YOLOv8n directly predicts the center of objects. The network output is passed through a Non-Maximum Suppression (NMS) filter, restricted to relevant COCO classes: \{2: Car, 3: Motorcycle, 5: Bus, 7: Truck\}. The network yields a set of continuous bounding boxes defined as $B = [x_1, y_1, x_2, y_2, conf\_score]$.

The composite loss function utilized during training is formulated as the sum of bounding box regression, objectness, and classification loss \cite{lin2017focal}:
\begin{equation}
L = L_{box} + L_{obj} + L_{cls}
\end{equation}
The regression loss ($L_{box}$) integrates Complete Intersection over Union (CIoU) alongside Distribution Focal Loss (DFL).

\subsection{Stage 2: Multi-Object Tracking (SORT)}
To associate independent vehicle detections across consecutive frames, SORT is applied \cite{bewley2016simple}. SORT models kinematics utilizing a discrete-time linear Kalman filter. The state of each tracked target is modeled as a seven-dimensional vector:
\begin{equation}
x = [u, v, s, r, \dot{u}, \dot{v}, \dot{s}]^T
\end{equation}
where $(u,v)$ represents the 2D pixel coordinate of the bounding box center, $s$ is the scale (pixel area), $r$ is the aspect ratio, and $\dot{u}, \dot{v}, \dot{s}$ represent respective velocities.

For sequential frame $k$, the Kalman filter predicts the \textit{a priori} state estimate $\hat{x}(k)$ based on the linear state transition matrix $F$ and the previous confirmed state $x(k-1)$:
\begin{equation}
\hat{x}(k) = F x(k-1) + w
\end{equation}
Data association is achieved via the Hungarian algorithm optimizing a bipartite graph. The assignment cost matrix utilizes Intersection over Union (IoU):
\begin{equation}
IoU(A,B) = \frac{|A \cap B|}{|A \cup B|}
\end{equation}

\subsection{Stage 3: License Plate Localization}
Upon successful tracking, the cropped region of interest (ROI) containing the vehicle is passed to a secondary, custom-trained YOLOv8 model (fine-tuned, 6.2 MB) dedicated exclusively to license plate localization. To prevent false associations, a rigid geometric assignment condition is enforced:
\begin{equation}
\begin{aligned}
plate_{x1} &> car_{x1} \land plate_{y1} > car_{y1} \land \\
plate_{x2} &< car_{x2} \land plate_{y2} < car_{y2}
\end{aligned}
\end{equation}

\subsection{Stage 4: OCR Processing Pipeline}
The localized plate ROI is subjected to an OCR pipeline driven by EasyOCR \cite{jaided2020easyocr} (English CPU mode). Prior to CRNN inference, the raw ROI undergoes deterministic pre-processing: BGR is converted to grayscale, followed by Binary Inverse Thresholding ($T=64$, $maxval=255$) utilizing OpenCV \cite{bradski2000opencv}.

The raw string is normalized (uppercase, stripped whitespace) and validated against the standardized 7-character UK license plate syntax: $[L][L][D][D][L][L][L]$. A positional character correction mapping table is applied:
\begin{itemize}
    \item \textbf{Expected Letters (Indices 0, 1, 4, 5, 6):} \{'0':'O', '1':'I', '3':'J', '4':'A', '6':'G', '5':'S'\}
    \item \textbf{Expected Digits (Indices 2, 3):} \{'O':'0', 'I':'1', 'J':'3', 'A':'4', 'G':'6', 'S':'5'\}
\end{itemize}

\subsection{Stage 5: Temporal Bounding Box Interpolation}
To address transient track losses resulting in discrete spatial gaps, a temporal interpolation algorithm is executed offline post-tracking. For each unique \texttt{car\_id}, missing bounding box coordinates $[x_1, y_1, x_2, y_2]$ are imputed using SciPy's \cite{virtanen2020scipy} 1-D interpolation function configured to a linear mode. The interpolated coordinate vector $C(t)$ at intermediate frame $t$ is defined as:
\begin{equation}
C(t) = C(t_0) + (t - t_0) \frac{C(t_1) - C(t_0)}{t_1 - t_0}
\end{equation}
Imputed rows are explicitly flagged with \texttt{license\_plate\_bbox\_score = 0} and \texttt{license\_number = 0} to preserve the integrity of downstream recognition metrics.

\section{Experimental Setup}
The experimental validation was performed by using a raw traffic surveillance video feed that only contains 3 599 continuous frames. The shots filmed cars going over in a multi-lane road, greatly exposed to natural fluctuation in day light, building shadows and uneven car speed.

It was implemented as the computational software stack based on Python 3.x and PyTorch to use the Ultralytics YOLOv8 models. The operations of image conversion were handled through the OpenCV 4.x. The process of tracking was implemented with the help of SORT, whereas the text transcription was provided with the assistance of EasyOCR. The data wrangling, as well as interpolation, were processed using Pandas, SciPy, and NumPy. On-purpose to induce edge-constrained situations, the EasyOCR module was constrained to only use CPU processing mode.

\section{Results and Analysis}
The processing of the 3,599-frame video sequence yielded a highly granular dataset. The base tracking algorithm generated the baseline dataset, while the post-processed, mathematically smoothed trajectory data formed the final interpolated dataset.

\begin{table}[htbp]
\caption{Dataset Summary}
\begin{center}
\begin{tabular}{lcc}
\toprule
\textbf{Metric} & \textbf{Baseline Dataset} & \textbf{Interpolated Dataset} \\
\midrule
Total entries & 2,247 & 4,536 \\
Imputed entries & - & 2,289 \\
Data increase & - & +101.9\% \\
Unique vehicle IDs & 45 & 45 \\
Total video frames & 3,599 & 3,599 \\
\bottomrule
\end{tabular}
\label{tab:dataset}
\end{center}
\end{table}

\begin{figure}[htbp]
\centerline{\includegraphics[width=\columnwidth]{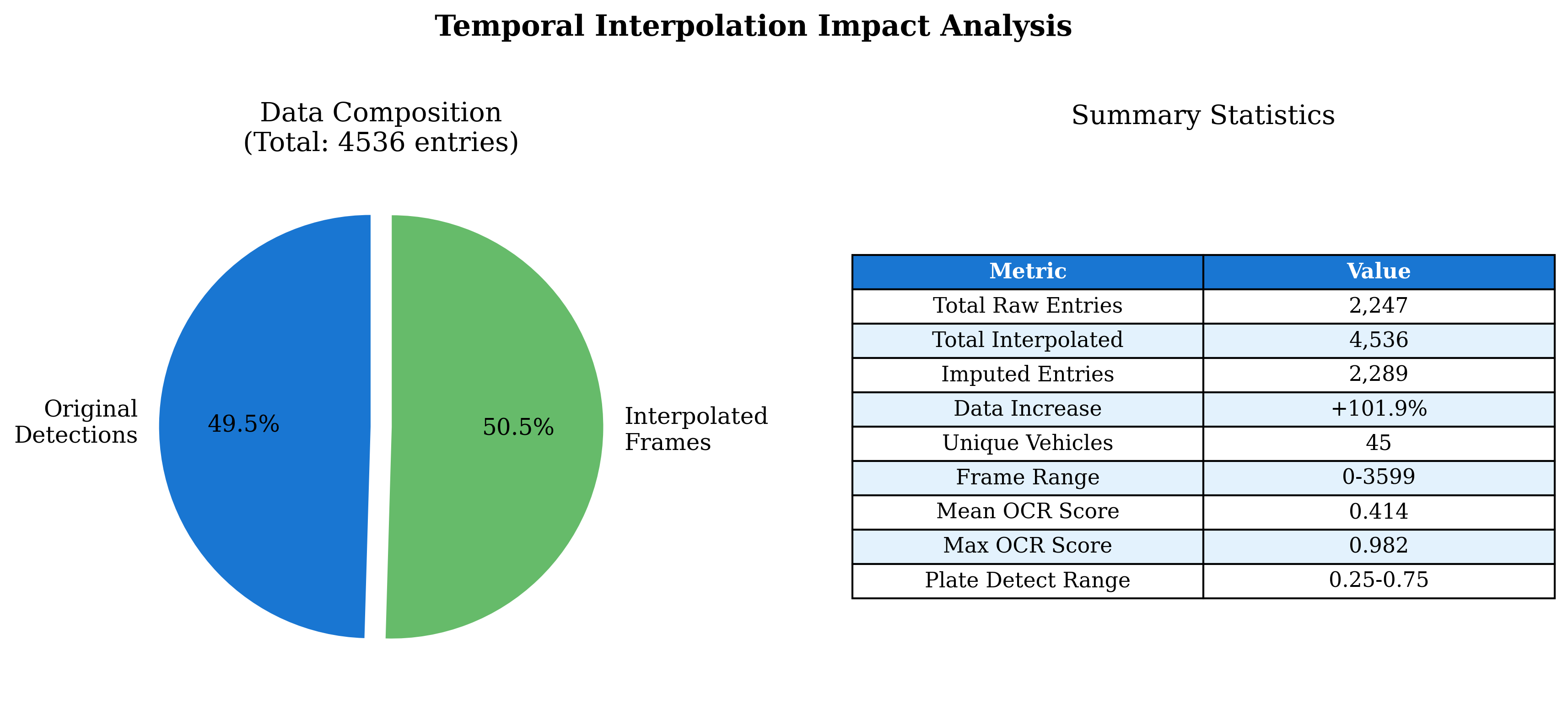}}
\caption{Dataset composition and summary statistics, demonstrating the impact of linear interpolation, which accounts for 50.5\% of the total entries.}
\label{fig:pie}
\end{figure}

As mentioned in Table \ref{tab:dataset} and visualized in Figure \ref{fig:pie}, the base pipeline yielded 2,247 discrete vehicle detections across 45 unique vehicle IDs. The application of linear temporal interpolation successfully generated 2,289 imputed rows, increasing the overall density of the spatial tracking data by 101.9\%.

\begin{figure}[htbp]
\centerline{\includegraphics[width=\columnwidth]{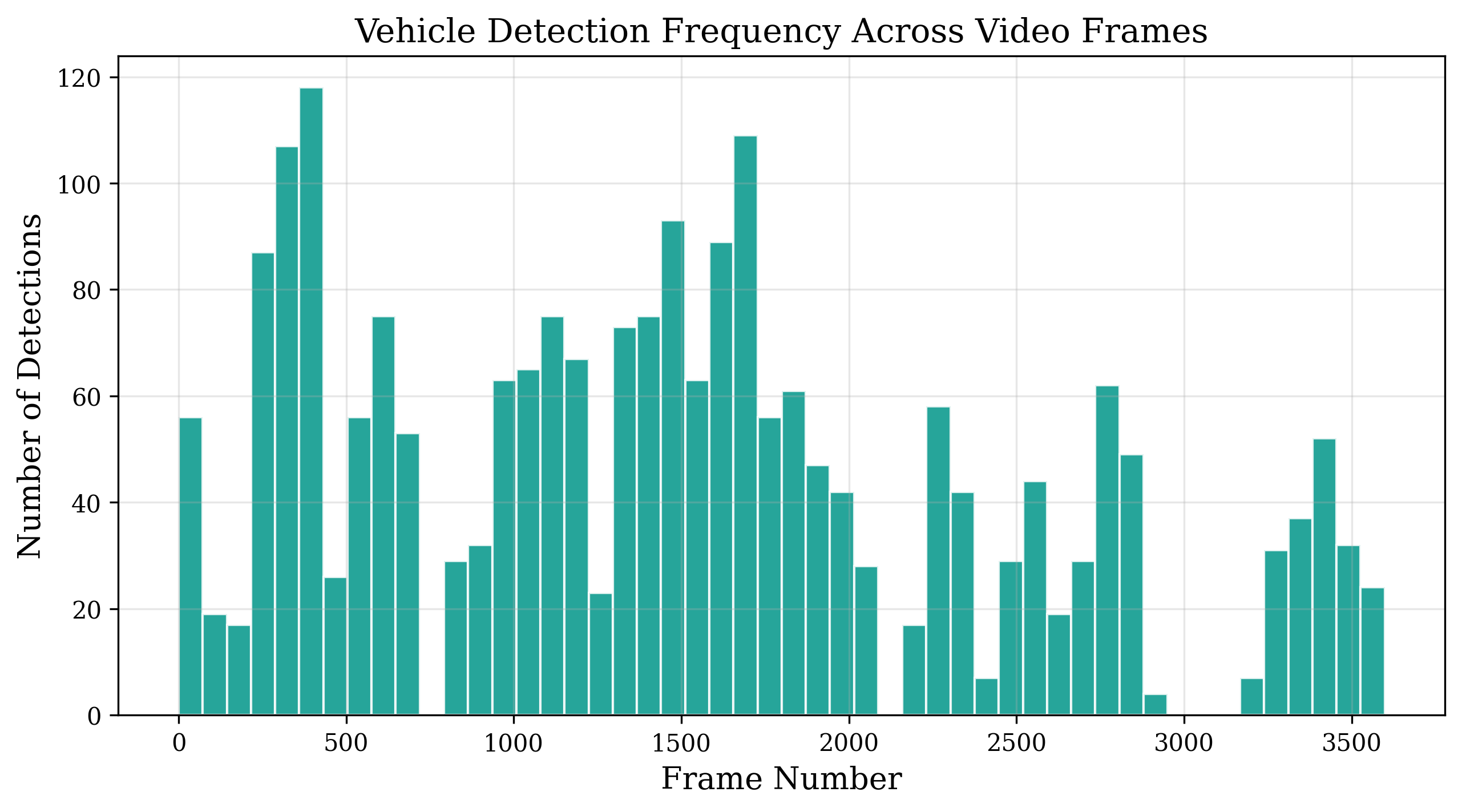}}
\caption{Vehicle detection frequency across the 3,599-frame video sequence, illustrating traffic density and flow dynamics over time.}
\label{fig:density}
\end{figure}

The temporal distribution of traffic flow is captured in Figure \ref{fig:density}, illustrating the frequency of vehicle detections across the video timeline, which aligns with intervals of high vehicle occlusion and dense traffic clustering.

\begin{figure}[htbp]
\centerline{\includegraphics[width=\columnwidth]{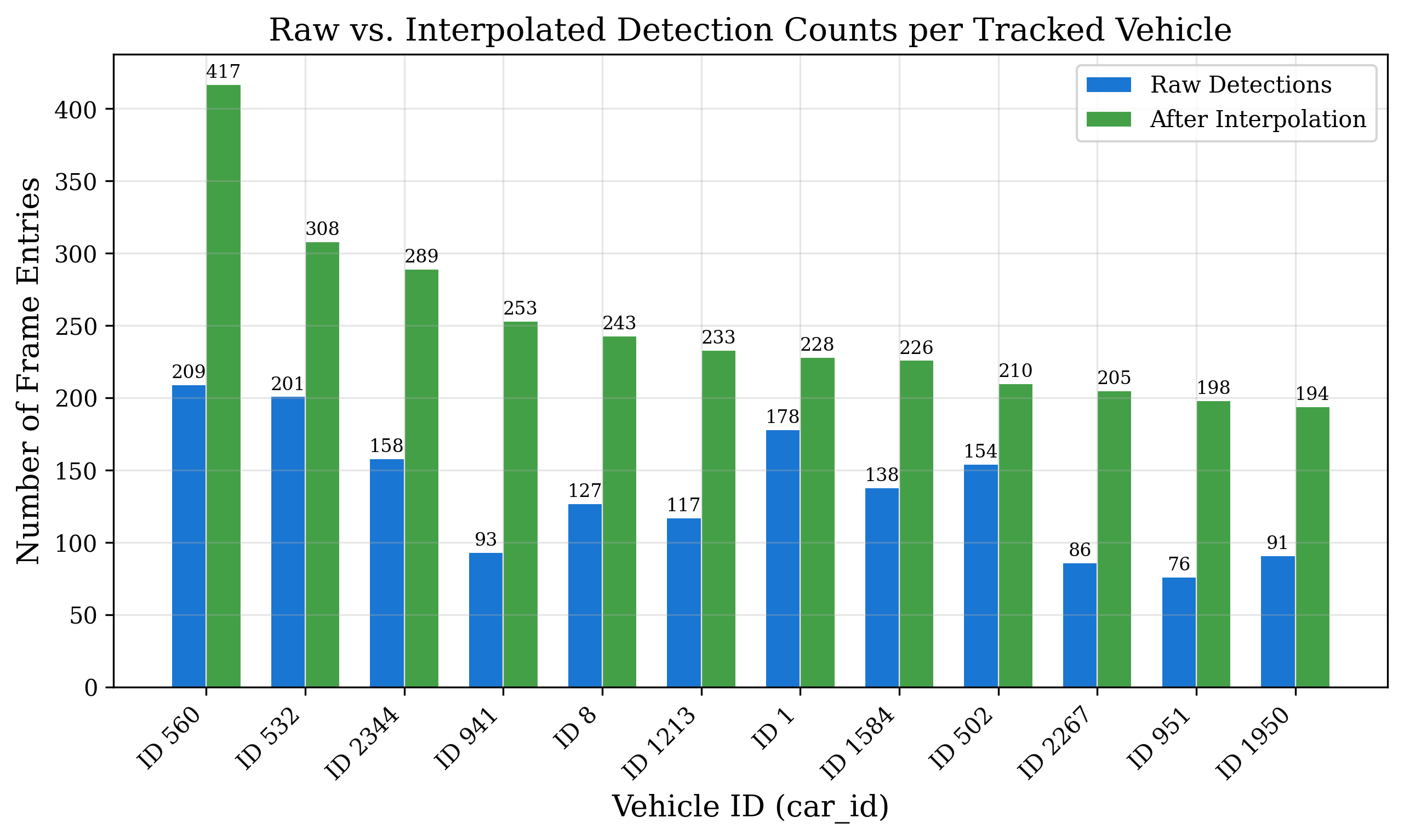}}
\caption{Grouped bar chart contrasting the baseline detection counts against the interpolated detection counts for tracked vehicles, showing massive data recovery.}
\label{fig:bar}
\end{figure}

\begin{figure}[htbp]
\centerline{\includegraphics[width=\columnwidth]{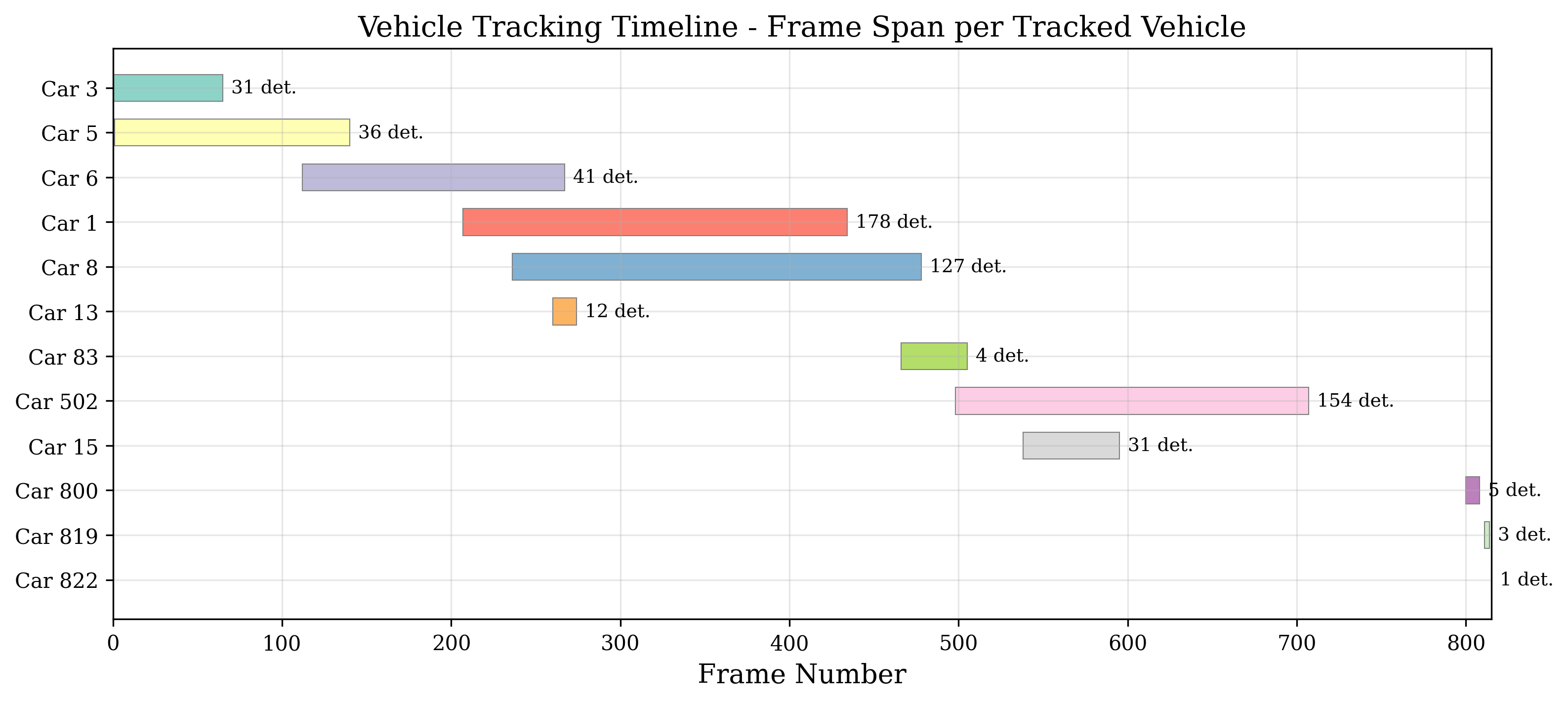}}
\caption{Horizontal Gantt-style timeline illustrating the presence and persistence of each vehicle across the frames, demonstrating continuous track recovery.}
\label{fig:gantt}
\end{figure}

Figure \ref{fig:bar} highlights the stabilization of trajectories per vehicle, showing how highly fragmented profiles exhibit massive gains post-interpolation. Figure \ref{fig:gantt} illustrates this persistence chronologically, charting the presence of each vehicle throughout the frame sequence.

\begin{table}[htbp]
\caption{OCR Statistical Summary}
\begin{center}
\begin{tabular}{lc}
\toprule
\textbf{Statistic} & \textbf{Value} \\
\midrule
Minimum & 0.0189 \\
Maximum & 0.9824 \\
Mean & 0.4140 \\
Median & 0.3942 \\
Std Deviation & 0.2078 \\
\bottomrule
\end{tabular}
\label{tab:ocr_stat}
\end{center}
\end{table}

\begin{figure}[htbp]
\centerline{\includegraphics[width=\columnwidth]{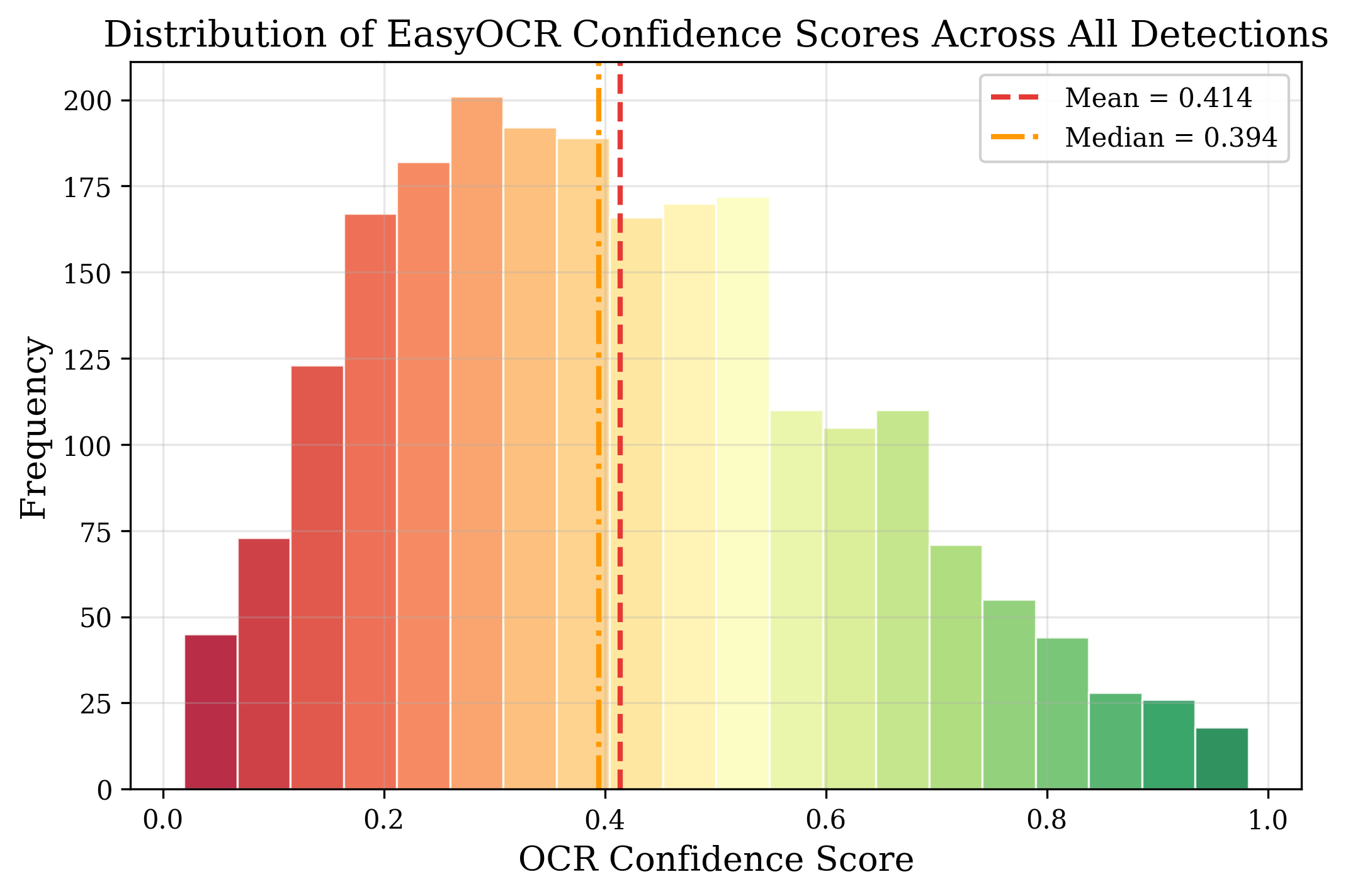}}
\caption{Density histogram of OCR confidence distribution, with mean and median markers highlighting the positive skew towards the lower-middle confidence tiers.}
\label{fig:hist}
\end{figure}

Table \ref{tab:ocr_stat} quantifies the performance of the EasyOCR engine. The mean OCR confidence across the dataset was 0.4140, with a significant standard deviation of 0.2078. Figure \ref{fig:hist} visually confirms this skew; only a small percentage of total reads achieved high confidence ($>0.80$).

\begin{figure}[htbp]
\centerline{\includegraphics[width=\columnwidth]{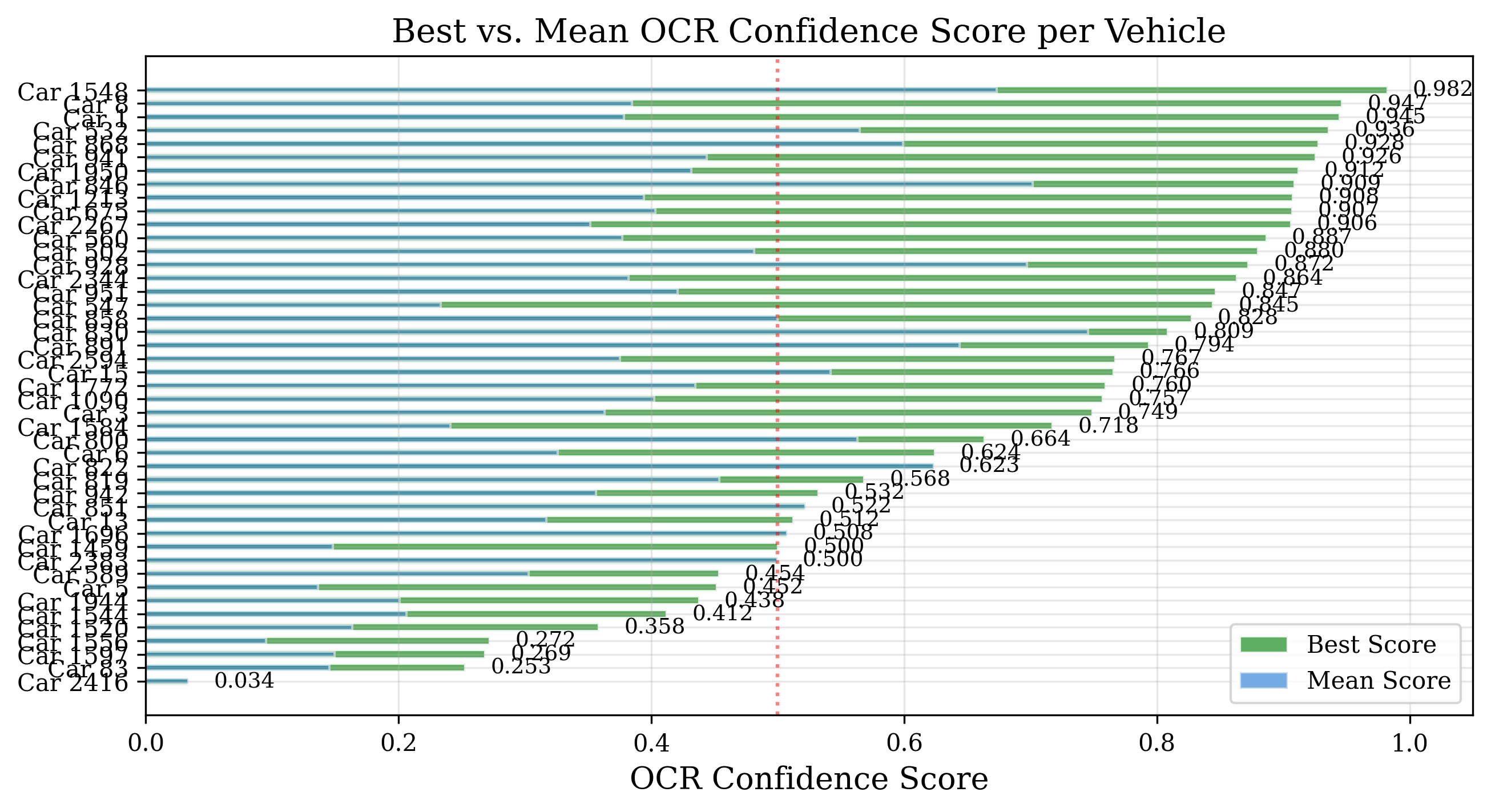}}
\caption{Horizontal bar chart comparing the peak absolute OCR score against the mean OCR score for each unique vehicle.}
\label{fig:best_vs_mean}
\end{figure}

\begin{figure}[htbp]
\centerline{\includegraphics[width=\columnwidth]{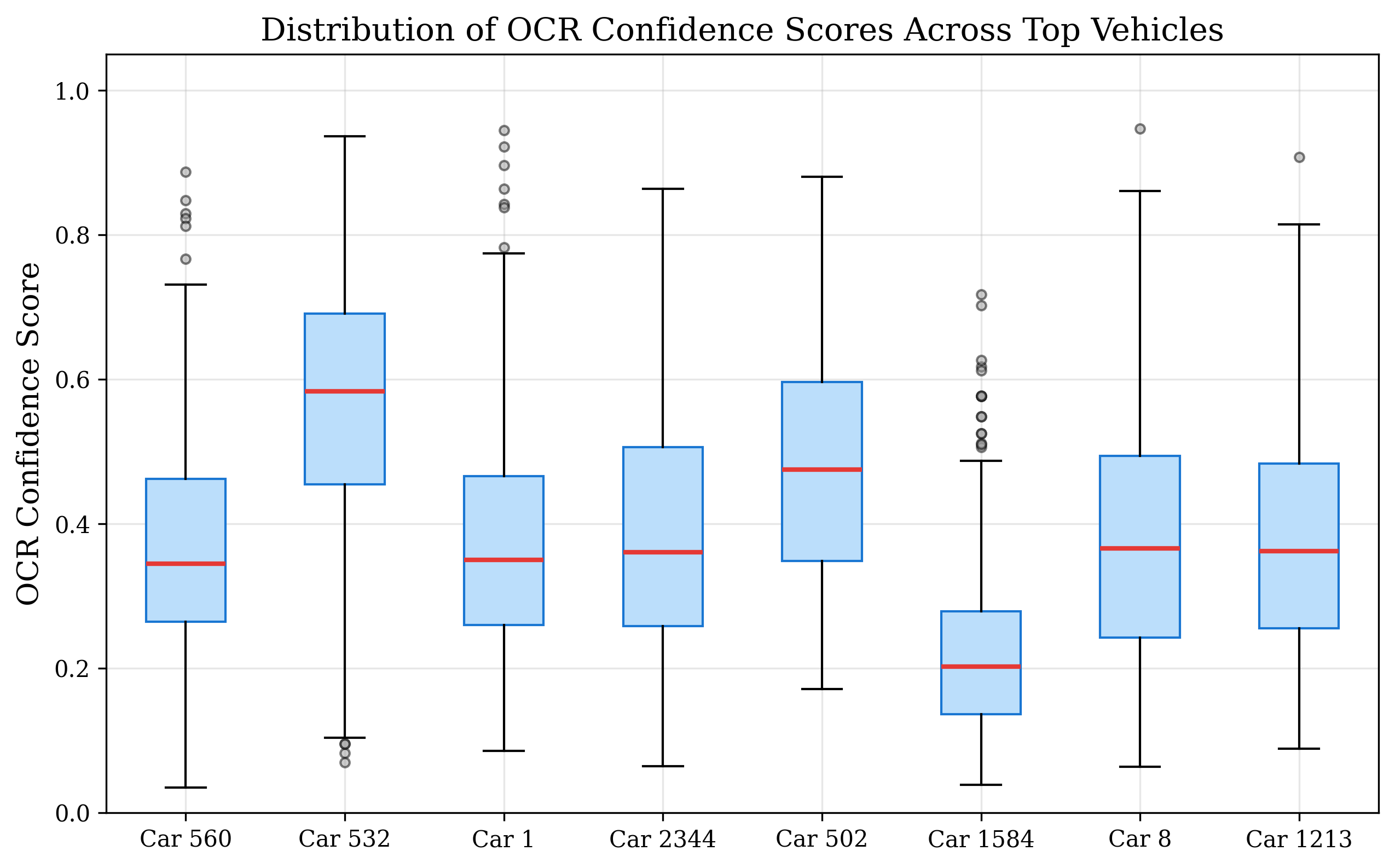}}
\caption{Distribution of OCR confidence scores across top vehicles, detailing interquartile ranges and performance disparities.}
\label{fig:boxplots}
\end{figure}

The severe intra-track variance is depicted in Figure \ref{fig:best_vs_mean}, contrasting peak vs. mean scores, and further emphasized in Figure \ref{fig:boxplots}, which provides a boxplot distribution highlighting median performance disparities across different physical vehicles.

\begin{figure}[htbp]
\centerline{\includegraphics[width=\columnwidth]{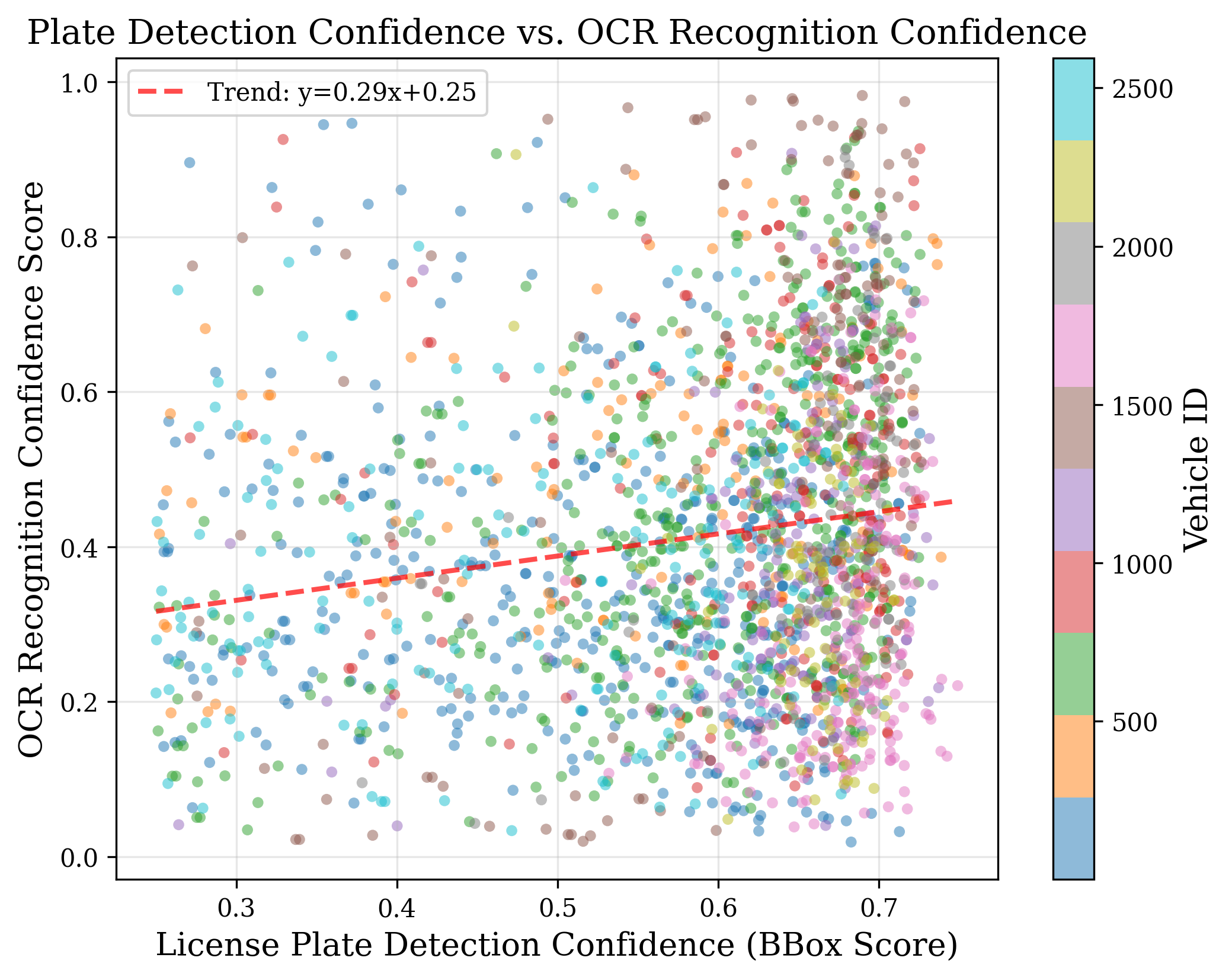}}
\caption{Scatter plot correlating YOLOv8 plate detection confidence against EasyOCR confidence, showing a positive but weak relationship.}
\label{fig:scatter_det_ocr}
\end{figure}

\begin{figure}[htbp]
\centerline{\includegraphics[width=\columnwidth]{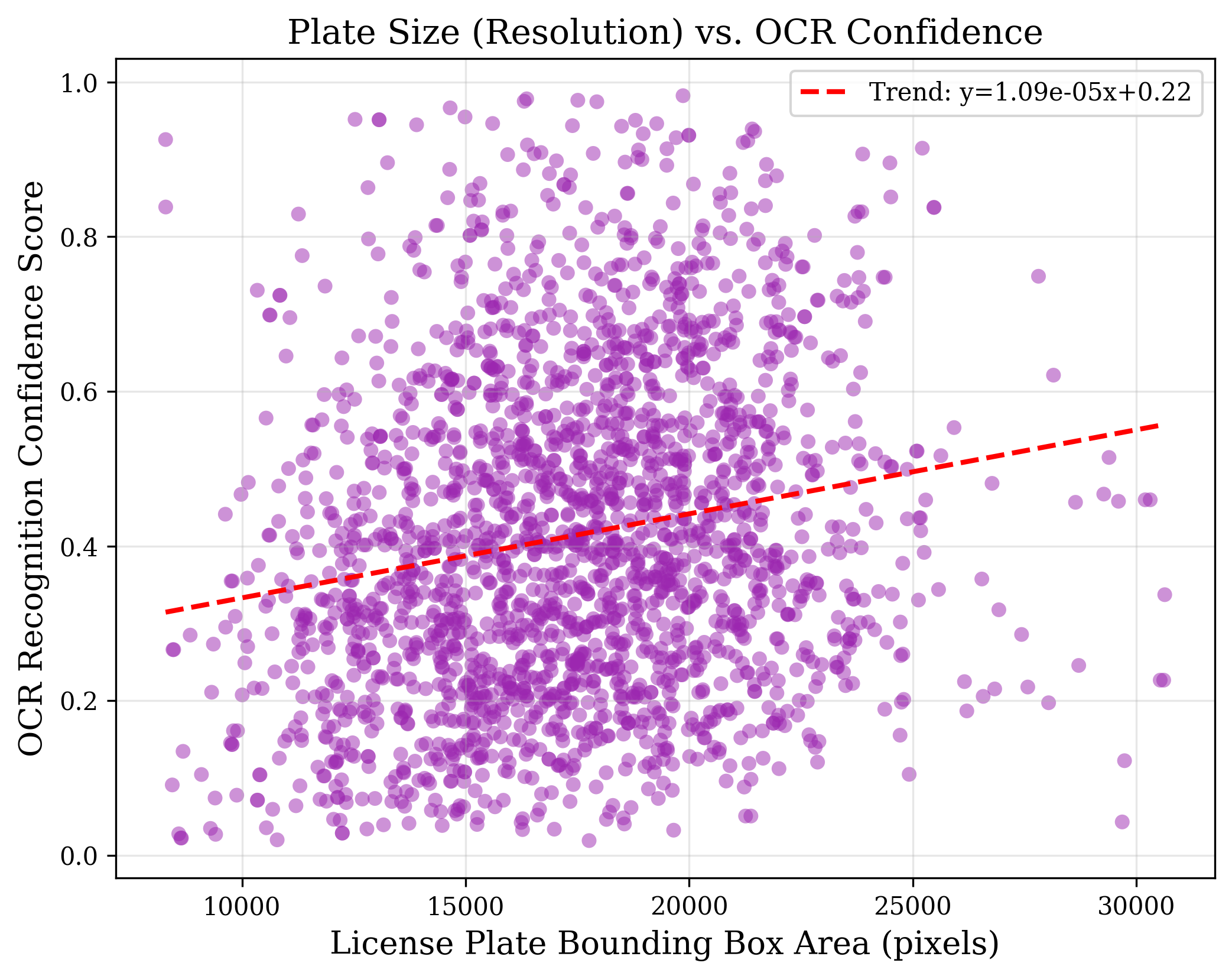}}
\caption{Scatter plot demonstrating the relationship between license plate bounding box area (resolution) and OCR recognition confidence.}
\label{fig:plate_area}
\end{figure}

Figure \ref{fig:scatter_det_ocr} maps plate detection confidence against OCR confidence. Furthermore, an analysis of plate bounding box area (resolution) against OCR confidence (Figure \ref{fig:plate_area}) demonstrates a positive correlation, indicating that larger, higher-resolution plate captures yield more reliable text extraction.

\begin{figure}[htbp]
\centerline{\includegraphics[width=\columnwidth]{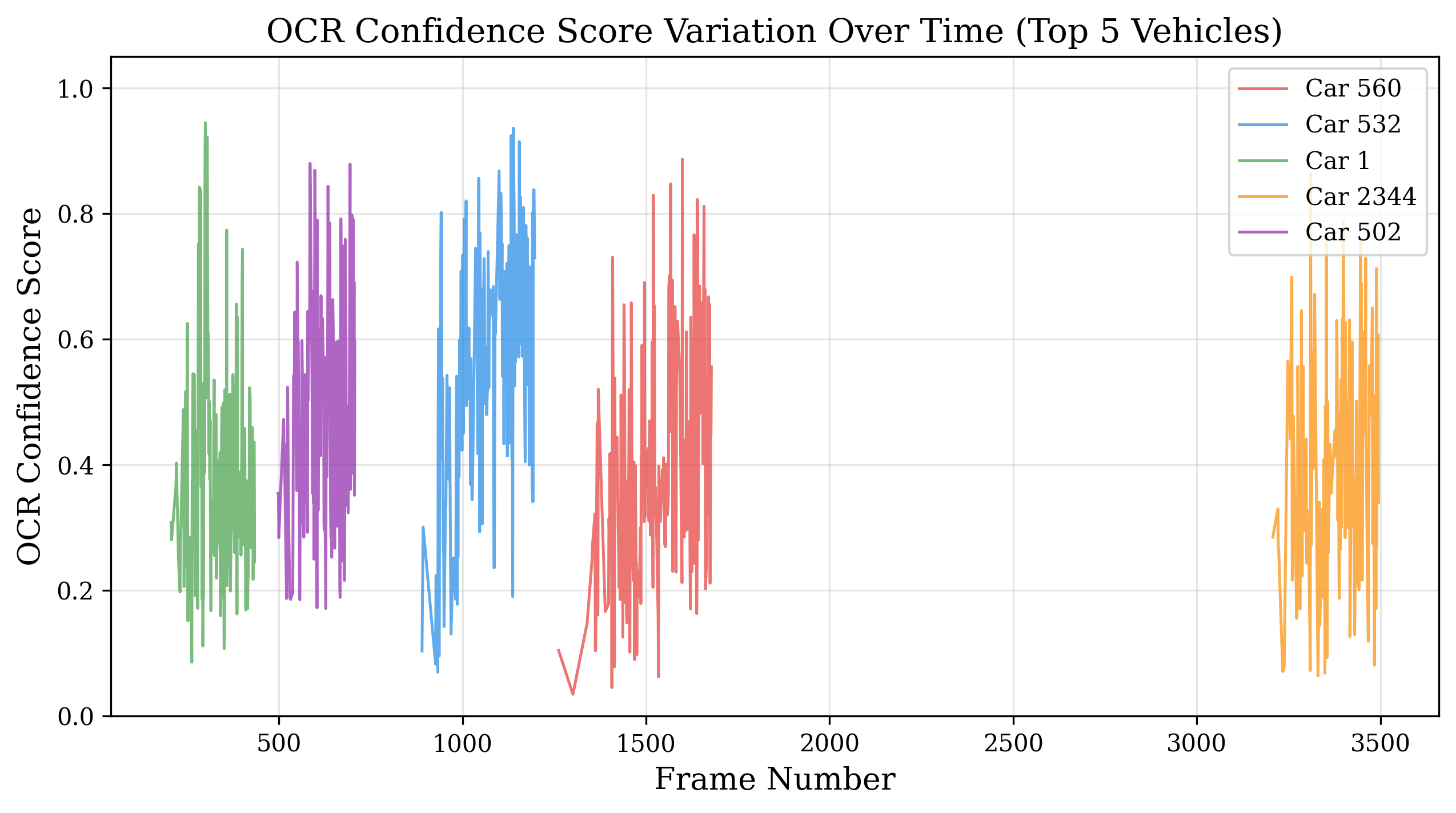}}
\caption{Line plot tracking OCR confidence over frame sequence for the top 5 most-detected vehicles, displaying extreme high-frequency oscillation.}
\label{fig:line}
\end{figure}

Figure \ref{fig:line} tracks OCR confidence over time for the top vehicles. The graph displays high-frequency oscillation, proving OCR confidence is highly transient and frame-dependent, rising sharply as vehicles approach the optimal focal plane.

\section{Discussion}
The practical evidence helps unravel deep information on the deployment functionalities of cascaded neural networks in ALPR. 

The chaired difference in the stability of OCR is one of the main analytical observations. Mean OCR confidence of 0.4140 shows that the extraction of text is extremely frame-dependent. Although YOLOv8 is able to localize boundaries with reasonable accuracy, motion blur corrupted internal pixel data, therefore, making feature illegible to the CRNN. The line in Figure \ref{fig:plate_area} illustrates that the density of the pixels (size of a plate) is paramount to the ability to overcome such environmental constraints.

It was mathematically essential to use positional character correction mapping. The post-processing module was able to intercept character ambiguities ( e.g. differences between I and 1) using the strict enforcement of the UK format. In the absence of this validation layer the raw EasyOCR output would have had a much higher rate of algorithm error based on font similarities.

Time data completion evaluation shows enormous quantitative worth in the choice of the trajectory smoothing. The linear interpolation added 2, 289 blank data to the spatial data and grew the spatial data by 101.9\% (Figure \ref{fig:pie}. Although such an interpolation willingly disables the text field of the occluded frames to avoid hallucination, it ensures that the coordinate trajectory of his/her downstream spatial analytics will be continuous. 

Nonetheless, even pure kinematic tracking paradigms such as SORT report structural limitations. Due to the use of the Kalman filter and bounding box IOU as the only two techniques, SORT experiences disastrous fragmentation in IDs in the absence of appearance-based feature encodings during extended occlusion.

\subsection{Limitations}
\begin{itemize}
    \item \textbf{Single Format Dependency:} The syntax correction is hardcoded exclusively to the UK format.
    \item \textbf{Computational Overhead:} CPU-bound EasyOCR inference severely bottlenecks the pipeline, precluding synchronous real-time deployment on standard edge devices.
    \item \textbf{Absence of Ground Truth:} Formal metrics (Precision, Recall, CER, WER) cannot be calculated without manually annotated ground truth bounding boxes.
    \item \textbf{Kinematic Assumptions:} Linear interpolation fails to accurately reconstruct bounding boxes for vehicles executing complex, non-linear maneuvers.
    \item \textbf{Moderate Accuracy Ceiling:} The aggregate mean OCR confidence is insufficient for fully automated law enforcement deployment, which requires thresholds exceeding 0.95.
\end{itemize}

\section{Data and Code Availability}
The full software pipeline, including the YOLOv8-SORT-EasyOCR integration scripts (\texttt{main.py}), temporal interpolation mechanisms (\texttt{add\_missing\_data.py}), and the array of data visualization generators used to produce the charts within this study (\texttt{generate\_graphs.py}, \texttt{visualize.py}), are publicly accessible and open-sourced. The repository can be found at \url{https://github.com/mobeen-pmo/Automatic-License-Plate-Recognition}.

\section{Conclusion}
This research successfully designed, implemented, and rigorously evaluated an end-to-end, five-stage Automatic License Plate Recognition pipeline, engineered by synthesizing YOLOv8n object detection, SORT multi-object tracking, and EasyOCR sequence transcription, fundamentally reinforced by a robust temporal data interpolation methodology. The proposed framework reduced temporal data loss in the face of extreme stochastics of environmental conditions of unrestrained traffic surveillance. The system was able to track 45 different vehicles out of a continuous 3,599 frames sequence.

The operational success of this architecture was the deepest with the interpolation of Stage 5 temporal bounding box. The linear imputation algorithm used recovered 2,289 missing data points by mathematically interpolating between spatial gaps due to occlusions. This achieved an absolute growth of 101.9\% in continuous trajectory coordinate, which was able to reduce fragmented point clouds to mathematically smooth vectors that can be used in more advanced ITS spatial analytics. More importantly, the inclusion of a deterministic dict of UK-style character correction turned out to be an indisputable necessity, which serves as some data boosting syntactic differentiating filter that smoothed the otherwise unstable CRNN outputs. 

Irrespective of such overtures, the thorough empirical investigations conclusively measured the definite shortcomings of lightweight, edge constrained aluminum probed ALPR arrays. The highest readings the system reached a confidence of 982 in text recognition, however, the combination of the average OCR ( 0.414) represents the instability of the optical character extraction when motion blur exists. Moreover, solely kinematic logic of MOT through SORT revealed serious weaknesses to elongated physical occlusions and induced extreme ID dispersal. Finally, this study offers a very thick, computationally elegant vindicated bottom design of trajectory-based ALPR, and effectively shows how algorithmic temporal interpolation is a vital integrative element of bowing intelligent traffic surveillances.

\end{document}